\documentclass[lettersize,journal]{IEEEtran}

\usepackage{amsmath,amssymb,amsfonts}
\usepackage{array}
\usepackage{xcolor}
\usepackage{graphicx}
\usepackage{booktabs}
\usepackage{multirow}
\usepackage{stfloats}
\usepackage{url}
\usepackage{cite}
\usepackage[lined,boxed,commentsnumbered,ruled]{algorithm2e}
\usepackage{algorithmic}
\begin{document}

\title{RiskFlow: Fast and Faithful Safety-Critical Traffic Scenario Generation}
\author{Jie Li, Qi Lan, Yining Tang, Yu Shen, Yi Zhou, Yuhao Wei, Guofa Li, Hao Sun
\thanks{\textit{(Corresponding authors: Guofa Li and Hao Sun.)}}
\thanks{J. Li, Q. Lan, Y. Tang, Y. Shen, Y. Zhou, Y. Wei, and G. Li are with the College of Mechanical and Vehicle Engineering, Chongqing University, Chongqing 400044, China.
{\tt\small Email: jieli@cqu.edu.cn; lanqi@stu.cqu.edu.cn;
tyn@stu.cqu.edu.cn; shenyu@stu.cqu.edu.cn;
yizhou@stu.cqu.edu.cn; yuhaowei@stu.cqu.edu.cn;
liguofa@cqu.edu.cn}}
\thanks{H. Sun is with Guangzhou Automobile Group Co., Ltd., Guangdong 511434, China.
{\tt\small Email: dr.sunhao@outlook.com}}}

\markboth{Submitted to IEEE Transactions on Intelligent Transportation Systems}{}

\maketitle

\begin{abstract}
Safety-critical traffic scenario generation is essential for evaluating autonomous driving systems under rare but high-risk interactions. Existing diffusion-based methods offer strong controllability in closed-loop generation, but their iterative denoising process is computationally expensive and may accumulate sampling and guidance errors over long rollouts, causing unrealistic motion artifacts such as abnormal acceleration and off-road behavior. To address these issues, we propose \textbf{RiskFlow}, a closed-loop safety-critical traffic generation framework that reformulates future behavior generation as one-step flow transport in the action space during training. At test time, RiskFlow generates safety-critical interactions through a selective guidance procedure in which time-to-collision-based critical-agent ranking selects high-risk agents and an adversarial interaction objective directly refines their output actions. Vehicle-dynamics rollout and map-aware regularization then translate guided actions into feasible trajectories while discouraging off-road behavior. By intervening on selected agents in the output space, RiskFlow avoids repeated corrections of intermediate denoising states, thereby mitigating accumulated guidance errors and accelerating closed-loop scenario generation. Experiments on the nuScenes dataset under closed-loop evaluation show that RiskFlow achieves a strong adversariality-realism trade-off across multi-agent and long-horizon settings. Compared with representative baselines, RiskFlow consistently improves realism while maintaining competitive safety-critical generation capability, and substantially reduces inference time for evaluation.
\end{abstract}

\begin{IEEEkeywords}
Autonomous driving, safety-critical scenario generation, modelling and simulation, flow matching
\end{IEEEkeywords}

\section{Introduction}

Autonomous driving systems should be evaluated not only in common traffic flows but also in rare safety-critical situations, such as collisions, forced cut-ins, emergency braking, and multi-vehicle merging. Collecting such events from real roads is costly and inefficient because crashes and severe safety outcomes are sparse in naturalistic driving logs and require enormous mileage to observe at scale~\cite{kalra2016driving}. Closed-loop safety-critical scenario generation has therefore become an important tool for stress-testing autonomous driving systems under controllable and repeatable high-risk interactions~\cite{ding2023survey}. The central challenge is to balance \emph{adversariality} and \emph{realism}: generated scenarios should induce challenging interactions while respecting road topology, traffic semantics, vehicle dynamics, and multi-agent behavior. Otherwise, high collision rates may result from excessive acceleration or off-road artifacts, providing limited value for safety evaluation.

Early studies improved the efficiency of automated-vehicle evaluation by increasing the occurrence probability of rare but critical events. Zhao et al.~\cite{zhao2016accelerated} used importance sampling to accelerate safety evaluation in lane-change cut-in scenarios while preserving the statistical characteristics of the original events. Feng et al.~\cite{feng2020testing} developed a testing scenario library generation framework in which scenario criticality is characterized by both exposure frequency and maneuver challenge. Subsequent studies explored surrogate-model-based gradient descent for identifying safety-performance boundaries~\cite{wang2022safety} and online adaptive generation of critical boundary scenarios~\cite{zhou2023online}. These approaches demonstrate the effectiveness of scenario-based evaluation, although they generally rely on predefined scenario parameterizations, application-specific evaluation models, or iterative optimization procedures.

\begin{figure}[t]
    \centering
    \includegraphics[width=0.95\columnwidth]{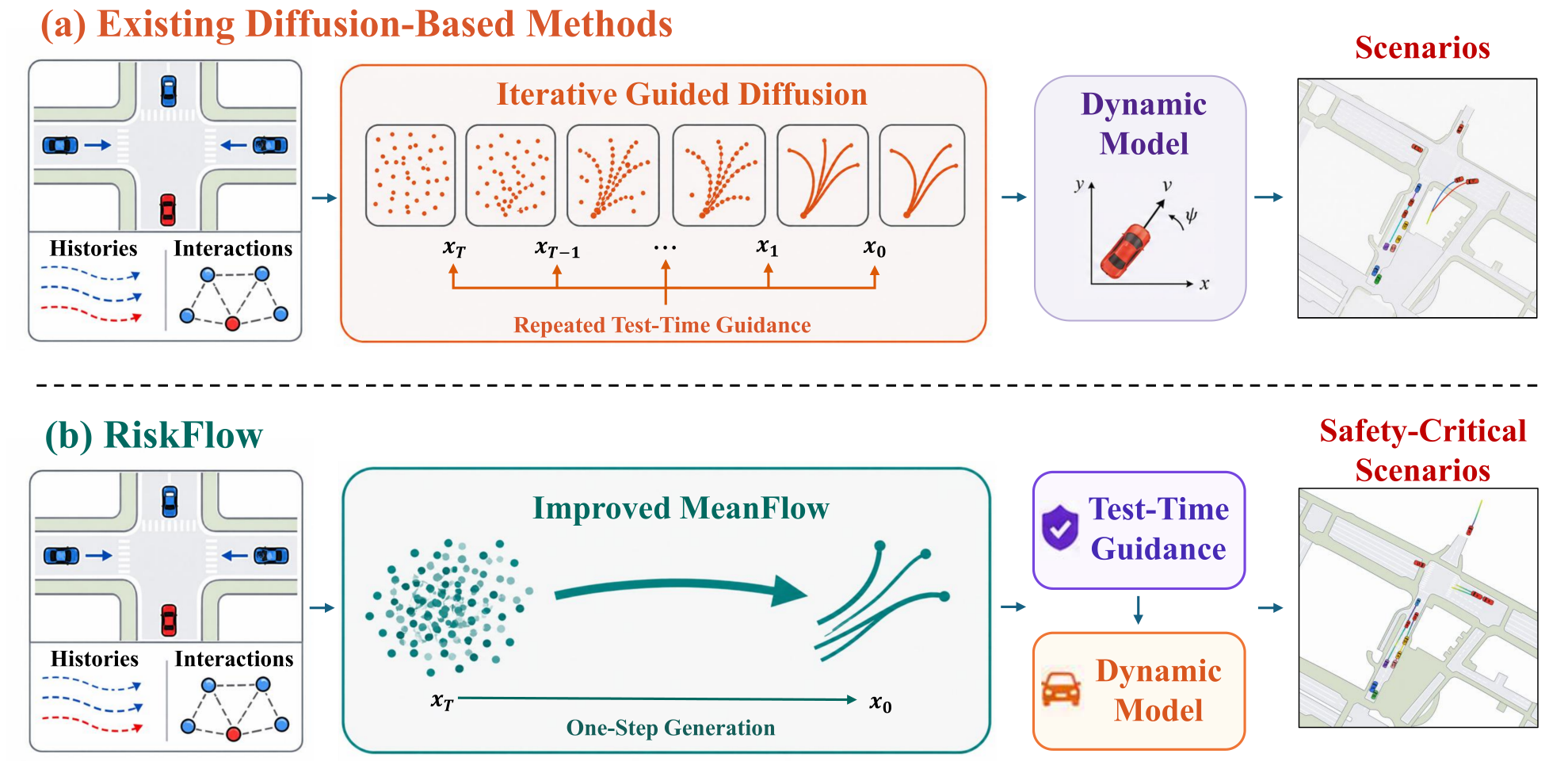}
    \caption{Comparison between existing diffusion-based methods and RiskFlow. (a) Existing methods rely on iterative guided diffusion. (b) RiskFlow enables one-step generation for efficient safety-critical scenario generation.}
    \label{fig:intro_comparison}
\end{figure}

Naturalistic traffic simulation improves closed-loop realism by learning interactive behavior models from large-scale datasets such as nuScenes~\cite{caesar2020nuscenes}. Methods such as SimNet, TrafficSim, and BITS roll out data-driven multi-agent behaviors that react to evolving scenes rather than replaying fixed trajectories~\cite{bergamini2021simnet,suo2021trafficsim,xu2023bits}. More recent joint-simulation frameworks explicitly model multimodal behaviors and reference paths for multiple interacting agents, improving collision avoidance, road compliance, and overall simulation realism~\cite{mo2025multimodal}. However, these simulators primarily aim to reproduce plausible traffic dynamics and maintain long-horizon rollout quality rather than deliberately induce safety-critical interactions.

Diffusion models have consequently become popular for controllable traffic generation because of their multimodal generative capability and flexible test-time guidance. Methods such as Scenario Diffusion~\cite{pronovost2023scenario}, CTG~\cite{zhong2023guided}, CTG++~\cite{zhong2023language}, SAFE-SIM~\cite{chang2024safe}, and CCDiff~\cite{lin2025causal} use conditional generation or differentiable guidance to produce diverse traffic scenes under safety-related objectives. VBD has combined multi-step diffusion refinement at inference time with behavior priors and model-based guidance to generate realistic and controllable multi-agent interactions, including long-tail safety-critical scenarios~\cite{huang2026versatile}. These developments indicate the potential of diffusion models for safety-critical traffic simulation but also highlight their reliance on iterative sampling.

Despite their effectiveness, diffusion-based methods have two key limitations, as illustrated in Fig.~\ref{fig:intro_comparison}. First, DDPM-style models require multi-step reverse denoising~\cite{ho2020denoising}. In closed-loop simulation, this sampling process must be executed repeatedly over successive planning cycles. For guided variants, safety or control objectives may also be evaluated multiple times during the refinement process. Sampling and guidance errors can consequently propagate through vehicle dynamics, particularly under strong collision-oriented guidance, leading to trajectory jitter, abnormal acceleration, shifted jerk distributions, or off-road behavior. Second, diffusion sampling is computationally expensive. Even accelerated samplers such as DDIM~\cite{song2020denoising} require repeated model evaluations, and this cost grows rapidly when many agents, scenes, and closed-loop planning cycles are considered. Efficient safety-critical generation therefore requires a mechanism that reduces both accumulated sampling error and inference cost.

Motivated by these observations, we propose \textbf{RiskFlow}, a flow-based framework for closed-loop safety-critical traffic generation. Given the initial traffic state, map context, and multi-agent history, RiskFlow starts from random action sequences and transforms them into future executable actions through a single MeanFlow forward pass in the action space. The generated acceleration and yaw-rate sequences are subsequently decoded into trajectories using vehicle dynamics. To induce safety-critical events, RiskFlow applies test-time guidance directly to the generated actions, steering selected agents toward risky interactions while discouraging off-road behavior through map-aware regularization. By refining the final actions rather than repeatedly modifying intermediate denoising states, RiskFlow can adjust safety-critical behaviors more directly while reducing trajectory distortion caused by accumulated sampling and guidance errors.

Our main contributions are summarized as follows:

\begin{itemize}
    \item We propose \textbf{RiskFlow}, a closed-loop safety-critical traffic scenario generation framework that formulates future action generation as a one-step flow transport problem. The model learns an average velocity field over future acceleration and yaw-rate sequences, allowing random noise to be transformed into executable action sequences through a single forward pass. To improve one-step reconstruction and training stability, we further adopt a JVP-based objective that evaluates the velocity field at the diagonal boundary $t=r$ and stops the gradient through the boundary velocity.

    \item We introduce test-time output-space guidance for single-pass generation. Instead of injecting guidance gradients throughout a multi-step denoising chain, RiskFlow directly refines the generated acceleration and yaw-rate sequences using an adversarial interaction objective and a trajectory-based map regularization term. This design enables safety-critical event generation while improving road feasibility and physical realism.

    \item Experiments on the nuScenes dataset demonstrate that RiskFlow achieves a strong trade-off between adversariality and realism in both multi-agent and long-horizon closed-loop generation while substantially reducing inference cost compared with diffusion-based baselines.
\end{itemize}

\section{Related Work}

\subsection{Traffic Behavior Modeling and Closed-Loop Simulation}

Existing traffic models support closed-loop evaluation through complementary simulation and prediction capabilities. SimNet captures reactions to ego actions~\cite{bergamini2021simnet}, TrafficSim models socially consistent joint behavior~\cite{suo2021trafficsim}, and BITS improves rollout stability through hierarchical behavior modeling~\cite{xu2023bits}. Multimodal joint simulation further improves collision avoidance and road compliance~\cite{mo2025multimodal}. Scene Transformer~\cite{ngiam2021scene}, QCNet~\cite{zhou2023query}, and Wayformer~\cite{nayakanti2023wayformer} encode agent histories, maps, and interactions, while causal forecasting improves robustness to distribution shifts~\cite{liu2022towards}. Counterfactual prediction~\cite{chen2021human} and intervention-based prediction~\cite{ge2023causal} reduce spurious correlations. Together, these methods strengthen realism and interaction modeling, but mainly serve as behavior priors or open-loop predictors rather than mechanisms for controllable safety-critical closed-loop generation.

\subsection{Safety-Critical Scenario Generation and Evaluation}

Scenario-based testing accelerates rare-event evaluation through probability reweighting and targeted search. Importance sampling increases exposure to critical lane-change cut-ins while preserving statistical characteristics~\cite{zhao2016accelerated}. Scenario libraries prioritize cases according to exposure frequency and maneuver challenge~\cite{feng2020testing}, whereas optimization-based search explores hazardous regions in logical parameter spaces~\cite{zhu2021hazardous}.

Response-aware approaches further adapt testing to vehicle performance. Surrogate-guided gradient descent identifies safety-performance boundaries~\cite{wang2022safety}. Online adaptive generation produces critical boundary scenarios through closed-loop feedback~\cite{zhou2023online}, while deep-RL-based adversarial evaluation learns diverse lane-change environments~\cite{chen2021adversarial}. Although efficient, these methods rely on predefined variables, task-specific evaluation models, iterative optimization, or separately trained policies. RiskFlow instead applies safety-critical objectives directly at inference without retraining its generative model.

\subsection{Diffusion-Based Scenario Generation}

Score-based modeling~\cite{song2019generative} and denoising diffusion~\cite{ho2020denoising} enable control by guiding iterative sampling~\cite{ho2022classifier}. Traffic applications span scene, motion, and traffic-state generation. Scenario Diffusion produces map-conditioned scenes~\cite{pronovost2023scenario}, MotionDiffuser incorporates constraints into multi-agent prediction~\cite{jiang2023motiondiffuser}, SceneDiffuser supports traffic simulation~\cite{jiang2024scenediffuser}, and ChatTraffic connects textual descriptions to traffic states~\cite{zhang2024chattraffic}.

Safety-critical methods increasingly structure guidance around rules, language, adversarial objectives, and interactions. CTG uses rule-based objectives~\cite{zhong2023guided} and CTG++ uses language instructions~\cite{zhong2023language}, while SAFE-SIM~\cite{chang2024safe} and AdvDiffuser~\cite{xie2024advdiffuser} guide adversarial agents. CCDiff encodes causal interactions~\cite{lin2025causal}, whereas VBD combines behavior priors with model-based refinement~\cite{huang2026versatile}. Conditional diffusion also generates high-risk transitions for policy learning~\cite{li2026reinforcement}. These methods provide flexible control but retain iterative denoising. RiskFlow instead generates actions in one step and applies adversarial and map-aware guidance in the action space.

\section{Method}

\begin{figure*}[t]
    \centering
    \includegraphics[width=0.95\textwidth]{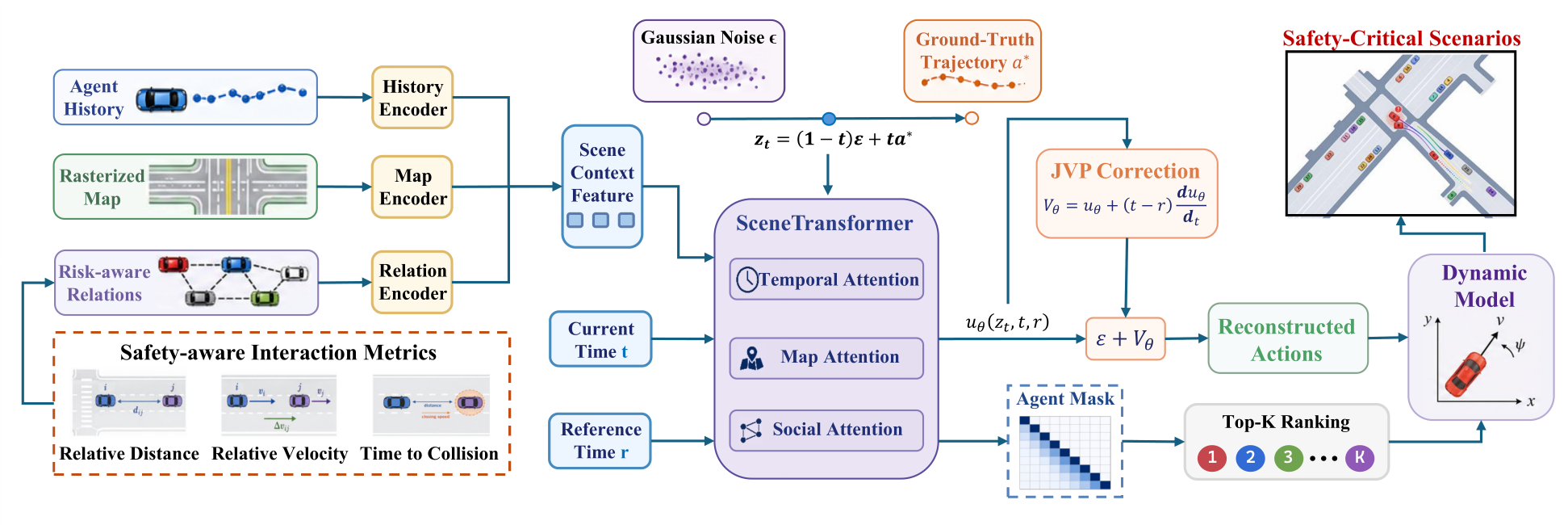}
    \caption{Overview of RiskFlow. RiskFlow encodes agent histories, rasterized map context, and risk-aware interaction features, then generates future actions with MeanFlow. At test time, TTC-based critical-agent ranking selects the top-$K$ high-risk agents for localized guidance.}
    \label{fig:RiskFlow}
\end{figure*}

In this section, we propose RiskFlow, a flow-based framework for closed-loop safety-critical traffic generation. As shown in
Fig.~\ref{fig:RiskFlow}, RiskFlow generates future action sequences from agent histories, rasterized maps, and risk-aware interaction features, and reconstructs physically feasible trajectories through vehicle dynamics.

\subsection{Problem Formulation}

Consider a traffic scene with $N$ vehicle agents. For each agent $i$, given its historical states, surrounding agents, and scene context, the goal is to generate a future action sequence over $L$ time steps in a closed-loop setting:
\begin{equation}
a_i^{1:L}
=
\{(\alpha_i^t,\omega_i^t)\}_{t=1}^{L},
\end{equation}
where $\alpha_i^t$ denotes longitudinal acceleration and $\omega_i^t$ denotes
yaw rate. The generated actions are integrated by a vehicle dynamics model to
obtain future states:
\begin{equation}
s_i^t
=
(x_i^t,y_i^t,v_i^t,\psi_i^t),
\end{equation}
where $(x_i^t,y_i^t)$ is the position, $v_i^t$ is the speed, and
$\psi_i^t$ is the heading angle at time step $t$. We use
$s_0=\{s_i^0\}_{i=1}^{N}$ to denote the current joint state that
initializes the vehicle dynamics rollout for subsequent closed-loop
action generation.

Rather than directly generating future positions, RiskFlow generates control
actions and reconstructs trajectories through a vehicle dynamics rollout. This action-space formulation provides a natural interface for closed-loop execution and encourages kinematic consistency in trajectories.

\subsection{Safety-Aware Scene Context Encoding}

To model safety-critical multi-agent interactions, we construct pairwise interaction features for each ordered agent pair $(i,j)$. Agent $j$ is represented in the local coordinate frame of agent $i$. Let $p_i^t,p_j^t\in\mathbb{R}^2$ denote the 2D
positions of agents $i$ and $j$ at time $t$, and let $\psi_i^t,\psi_j^t$ denote their heading angles. The relative position is computed as
\begin{equation}
\Delta p_{ij}^t
=
(R_i^t)^\top(p_j^t-p_i^t),
\end{equation}
where $R_i^t$ is the rotation matrix determined by $\psi_i^t$.

We represent the pairwise interaction feature as
\begin{equation}
\begin{aligned}
e_{ij}^t = [&
\Delta x_{ij}^t,
\Delta y_{ij}^t,
\cos \Delta\psi_{ij}^t,
\sin \Delta\psi_{ij}^t, \\
&
\Delta v_{x,ij}^t,
\Delta v_{y,ij}^t,
\tilde d_{ij}^t,
r_{ij}^t
],
\end{aligned}
\end{equation}
where $\tilde d_{ij}^t$ is a size-aware normalized distance feature and $r_{ij}^t$ is a normalized TTC-based risk weight. The TTC risk is estimated from relative position, relative velocity, and vehicle extents. We compute directional collision times along the longitudinal and lateral axes and define
\begin{equation}
\mathrm{TTC}_{ij}^t
=
\max(\tau_{x,ij}^t,\tau_{y,ij}^t).
\end{equation}
The TTC value is clipped and normalized into a risk weight:
\begin{equation}
r_{ij}^t
=
\frac{
\tau_{\max}
-
\mathrm{clip}(\mathrm{TTC}_{ij}^t,0,\tau_{\max})
}{
\tau_{\max}
}.
\end{equation}
A larger $r_{ij}^t$ indicates a higher short-term collision risk under the current relative motion.

We use a scene transformer backbone to encode agent histories, pairwise interactions, and rasterized map context~\cite{ngiam2021scene}. Temporal attention captures each agent's motion pattern, while social attention models inter-agent interactions. To incorporate safety-aware information, we encode each pairwise feature and concatenate it with the neighboring agent representation:
\begin{equation}
g_{ij}=f_{\mathrm{edge}}(e_{ij}),
\qquad
\tilde h_{ij}=[h_j\,\|\,g_{ij}].
\end{equation}
The edge-conditioned query, key, and value are defined as
\begin{equation}
q_i=W_Qh_i,\qquad
k_{ij}=W_K\tilde h_{ij},\qquad
v_{ij}=W_V\tilde h_{ij}.
\end{equation}
Social attention is then computed as
\begin{equation}
\mathrm{Attn}_i
=
\sum_{j=1}^{N}
\operatorname{softmax}_{j}
\left(
\frac{q_i^\top k_{ij}}{\sqrt{d}}
\right)v_{ij}.
\end{equation}
This allows relative motion, vehicle geometry, and TTC-based risk to condition multi-agent interaction modeling.

Rasterized maps provide static road context for road-aware generation. For each agent-centric scene, a CNN map encoder extracts a local map feature
\begin{equation}
m_i = f_{\mathrm{map}}(I_i),
\end{equation}
where $I_i$ is the rasterized map centered at agent $i$ in its local frame, and $m_i$ is the corresponding encoded map feature. The conditional scene context is constructed as
\begin{equation}
C_i
=
f_{\mathrm{ctx}}
\left(
h_i,
m_i,
\{e_{ij}\}_{j=1}^{N}
\right),
\end{equation}
where $f_{\mathrm{ctx}}$ denotes the scene-context encoder that fuses the agent history, map, and safety-aware pairwise interaction features.
We collect all agent-level contexts into the joint scene context
$C=\{C_i\}_{i=1}^{N}$. RiskFlow then conditions the action-space
velocity field on $C$ during future action generation.

\subsection{MeanFlow Generation}

Instead of iterative reverse diffusion, RiskFlow adopts MeanFlow in the future action space~\cite{geng2026mean}. Let $a^\ast$ denote the normalized ground-truth future action sequence and let $\epsilon\sim\mathcal{N}(0,\mathbf{I})$ be Gaussian noise of the same shape.
We construct the interpolation

\begin{equation}
z_t=(1-t)\epsilon+t a^\ast,
\qquad t\in[0,1].
\end{equation}

The mean velocity field $u_\theta(z_t,t,r;C)$ predicts the context-conditioned average velocity from current time $t$ to reference time $r$~\cite{geng2025improved}. At the boundary $r=t$, we evaluate
$v_t=u_\theta(z_t,t,t;C)$ and compute the Jacobian-vector product:
\begin{equation}
\mathcal{J}_\theta
=
\operatorname{JVP}_{(z,t)}
\!\left[u_\theta(z,t,r;C)\right]_{(z_t,t)}
(v_t,\mathbf{1}).
\label{eq:jvp_correction}
\end{equation}
The corresponding training-time instantaneous velocity is

\begin{equation}
\tilde v_\theta
=
u_\theta(z_t,t,r;C)
+
(t-r)\operatorname{clip}
\left(
\operatorname{sg}(\mathcal{J}_\theta),-c,c
\right),
\end{equation}
where $\operatorname{sg}(\cdot)$ denotes stop-gradient and $c$ is the JVP clipping threshold. The field $\tilde v_\theta$ is used only to construct the training objective, whereas $u_\theta$ remains the average velocity used for one-step base action generation.

The reconstructed actions and resulting states are

\begin{equation}
\hat a=\epsilon+\tilde v_\theta,
\qquad
\hat s^{1:L}=F_{\mathrm{dyn}}(\hat a,s_0).
\end{equation}
Let $\hat y^\ell=[\hat s^\ell,\hat a^\ell]$ and
$y^{\ast,\ell}=[s^{\ast,\ell},a^{\ast,\ell}]$ denote the normalized predicted and ground-truth state-action vectors at future step $\ell$.
We optimize the weighted reconstruction objective

\begin{equation}
\mathcal{L}_{\mathrm{MF}}
=
\mathbb{E}_{a^\ast,\epsilon,t,r}
\left[
\frac{1}{|\mathcal{V}|D}
\sum_{\ell\in\mathcal{V}}
\sum_{d=1}^{D}
w_d
\left(
\hat y_d^\ell-y_d^{\ast,\ell}
\right)^2
\right],
\end{equation}
where $\mathcal{V}$ contains the valid future steps, $D$ is the
state-action dimension, and $w_d$ is the reconstruction weight of dimension $d$. The action component supervises
$\tilde v_\theta$ toward $a^\ast-\epsilon$, while the state component encourages dynamically consistent trajectories.

The complete training procedure, including action interpolation,
JVP-corrected velocity estimation, dynamics rollout, and parameter
optimization, is summarized in
Algorithm~\ref{alg:riskflow_training}.

\begin{algorithm}[!tb]
\caption{MeanFlow-Based Training of RiskFlow}
\label{alg:riskflow_training}
\LinesNumbered

\KwIn{Histories $H=\{H_i\}_{i=1}^{N}$ and maps
$\mathcal{I}=\{I_i\}_{i=1}^{N}$; initial states $s_0$;
target states $s^\ast$ and normalized target actions $a^\ast$;
valid future indices $\mathcal{V}$; normalizers
$\mathcal{N}_a$ and $\mathcal{N}_y$; reconstruction weights
$\{w_d\}_{d=1}^{D}$; JVP clipping threshold $c$;
training steps $N_{\mathrm{train}}$; trainable parameters
$\theta$; fixed dynamics $F_{\mathrm{dyn}}$;
learning rate $\eta$.}

Map the target actions to physical units
$a_{\mathrm{phy}}^\ast
\gets\mathcal{N}_a^{-1}(a^\ast)$\;

\For{$m\gets1$ \KwTo $N_{\mathrm{train}}$}{
    Construct the pairwise interaction features $e_{ij}$ and
    TTC-based risk features $r_{ij}$\;

    Encode the conditional scene representation
    $C\gets f_\theta
    (H,\mathcal{I},\{e_{ij},r_{ij}\})$\;

    Sample Gaussian action noise
    $\epsilon\sim\mathcal{N}(0,\mathbf{I})$\;

    Sample $(t,r)$ from the logit-normal time-pair
    distribution, where $0\leq t\leq r\leq1$\;

    Construct the interpolated action sequence
    $z_t\gets(1-t)\epsilon+t a^\ast$\;

    Evaluate the diagonal velocity
    $v_t\gets u_\theta(z_t,t,t;C)$\;

    Compute the Jacobian-vector product
    $\mathcal{J}_\theta\gets
    \operatorname{JVP}_{(z,t)}
    [u_\theta(z,t,r;C)]_{(z_t,t)}
    (v_t,\mathbf{1})$\;

    Construct the corrected instantaneous velocity
    $\tilde v_\theta\gets
    u_\theta(z_t,t,r;C)
    +(t-r)\operatorname{clip}
    (\operatorname{sg}(\mathcal{J}_\theta),-c,c)$\;

    Reconstruct the normalized future actions
    $\hat a_{\mathrm{n}}
    \gets\epsilon+\tilde v_\theta$\;

    Map the predicted actions to physical units
    $\hat a\gets
    \mathcal{N}_a^{-1}(\hat a_{\mathrm{n}})$\;

    Roll out the corresponding future states
    $\hat s^{1:L}\gets
    F_{\mathrm{dyn}}(\hat a,s_0)$\;

    Construct the normalized predicted vectors
    $\hat y^\ell\gets
    \mathcal{N}_y([\hat s^\ell,\hat a^\ell])$\;

    Construct the normalized target vectors
    $y^{\ast,\ell}\gets
    \mathcal{N}_y
    ([s^{\ast,\ell},a_{\mathrm{phy}}^{\ast,\ell}])$\;

    Compute the weighted reconstruction loss
    $\mathcal{L}_{\mathrm{MF}}\gets
    \mathbb{E}_{a^\ast,\epsilon,t,r}
    \left[
    \frac{1}{|\mathcal{V}|D}
    \sum_{\ell\in\mathcal{V}}
    \sum_{d=1}^{D}
    w_d
    \left(
    \hat y_d^\ell-y_d^{\ast,\ell}
    \right)^2
    \right]$\;

    Update $\theta$ using Adam with learning rate $\eta$\;
}

\KwRet{Updated model parameters $\theta$}
\end{algorithm}

\subsection{Output-Space Test-Time Guidance}

At planning cycle $q$, given current state $s^q$ ($s^0=s_0$) and
updated scene context $C^q$, RiskFlow obtains the base action residual by evaluating the learned average velocity field once:
\begin{equation}
u_{\mathrm{base}}^{(0)}
=
u_\theta(\epsilon,0,1;C^q),
\qquad
a_{\mathrm{base}}
=
\epsilon+u_{\mathrm{base}}^{(0)}.
\end{equation}

Unlike the training-time field $\tilde v_\theta$, this one-step base prediction requires no JVP computation. RiskFlow then refines the base residual through output-space guidance. At guidance step $k$, the corresponding actions and trajectories are
\begin{equation}
a^{(k)}
=
\epsilon+u_{\mathrm{base}}^{(k)},
\qquad
\tau^{(k)}
=
F_{\mathrm{dyn}}(a^{(k)},s^q).
\end{equation}

Let $\hat s_i^{\ell,(k)}$ and $\hat p_i^{\ell,(k)}$ denote agent $i$'s predicted state and position at future step $\ell$ in guided trajectory $\tau^{(k)}$. 
Let $\mathcal{A}$ denote the set of all valid agents in the current scene, and let $A_c\subseteq\mathcal{A}$ denote the selected controllable agent set. When at least one non-controlled target is available, the adversarial interaction objective is
\begin{equation}
\mathcal{L}_{\mathrm{adv}}
=
\sum_{i\in A_c}
\sum_{\ell=1}^{L}
\min_{j\in\mathcal{A}\setminus A_c}
\left\|
\hat p_i^{\ell,(k)}
-
\hat p_j^{\ell,(k)}
\right\|_2 .
\end{equation}
Minimizing this objective encourages controllable agents to approach nearby non-controlled agents, producing challenging high-risk interactions.

Optimizing only the adversarial interaction objective may generate collisions through unrealistic lateral shifts or off-road motion. To preserve road feasibility, we introduce a footprint-based map guidance term:
\begin{equation}
\mathcal{L}_{\mathrm{map}}
=
\sum_{i\in A_c}
\sum_{\ell=1}^{L}
\omega_\ell
\Phi_{\mathrm{map}}
\left(
\hat s_i^{\ell,(k)}
\right),
\end{equation}
where $\Phi_{\mathrm{map}}$ penalizes vehicle footprint samples that leave the drivable region, and $\omega_\ell$ is a temporal decay weight. This term discourages off-road artifacts while allowing the adversarial objective to induce high-risk interactions.

The final guidance objective combines the two terms:
\begin{equation}
\mathcal{L}_{\mathrm{guide}}
=
\lambda_{\mathrm{adv}}\mathcal{L}_{\mathrm{adv}}
+
\lambda_{\mathrm{map}}\mathcal{L}_{\mathrm{map}},
\end{equation}
where $\lambda_{\mathrm{adv}}$ and $\lambda_{\mathrm{map}}$ control the strengths of adversarial interaction guidance and map-feasibility guidance.

To localize the intervention, a binary mask $M_{A_c}$ restricts residual updates to agents in $A_c$. At each guidance step, Adam refines the residual:
\begin{equation}
u_{\mathrm{base}}^{(k+1)}
=
\operatorname{Adam}_{\gamma}
\left(
u_{\mathrm{base}}^{(k)},
M_{A_c}\odot
\nabla_{u_{\mathrm{base}}^{(k)}}
\mathcal{L}_{\mathrm{guide}}
\right),
\end{equation}
where $\operatorname{Adam}_{\gamma}$ denotes one Adam update with learning rate $\gamma$ for minimizing the guidance objective. After $K_g$ guidance steps, the final action
sequence is
\begin{equation}
a_{\mathrm{final}}
=
\epsilon+u_{\mathrm{base}}^{(K_g)},
\end{equation}
where $K_g$ is the number of guidance steps.

\subsection{Closed-Loop Generation Procedure}

Algorithm~\ref{alg:riskflow_inference} summarizes the closed-loop generation procedure of RiskFlow. Before each rollout, the top-$K$ controllable agents are selected according to a precomputed TTC-based critical-agent ranking and remain fixed throughout the rollout. At each planning cycle, a single evaluation of $u_\theta$ at $(t,r)=(0,1)$ transforms Gaussian noise into a base action residual. The residual is then refined through $K_g$ output-space guidance steps without additional evaluations of $u_\theta$. Each guidance step performs a differentiable dynamics rollout, evaluates the adversarial and map-feasibility objectives, and masks the resulting gradient to the selected agents. After optimization, the first $L_e$ actions are executed, and RiskFlow replans from the updated state and history.This procedure repeats until the prescribed simulation horizon is reached. Therefore, one-step generation refers to the base-network prediction, whereas the complete closed-loop procedure additionally includes output-space guidance and vehicle-dynamics execution. 

\begin{algorithm}[!tb]
\caption{Closed-Loop Generation With RiskFlow}
\label{alg:riskflow_inference}
\LinesNumbered

\KwIn{Initial state $s^0$, history $H^0$, and map
$I_{\mathrm{map}}$; trained vector field $u_\theta$;
action normalizer $\mathcal{N}_a$; vehicle dynamics
$F_{\mathrm{dyn}}$; TTC-based critical-agent ranking $\pi$;
parameters $K$, $K_g$, $\gamma$, $L_e$, $Q$,
$\lambda_{\mathrm{adv}}$, and $\lambda_{\mathrm{map}}$.}

\KwOut{Closed-loop trajectory $\mathcal{T}$.}

Select the controlled agents
$A_c\gets\operatorname{FirstK}(\pi,K)$\;

Initialize the closed-loop trajectory
$\mathcal{T}\gets\{s^0\}$\;

\For{$q\gets0$ \KwTo $Q-1$}{
    Construct the scene context $C^q$\;

    Sample Gaussian action noise
    $\epsilon\sim\mathcal{N}(0,\mathbf{I})$\;

    Generate the normalized base action residual
    $u_{\mathrm{base}}^{(0)}
    \gets u_\theta(\epsilon,0,1;C^q)$\;

    \For{$k\gets0$ \KwTo $K_g-1$}{
        Reconstruct the normalized action sequence
        $a_{\mathrm{n}}^{(k)}
        \gets\epsilon+u_{\mathrm{base}}^{(k)}$\;

        Map the actions to physical units
        $a^{(k)}
        \gets\mathcal{N}_a^{-1}(a_{\mathrm{n}}^{(k)})$\;

        Roll out the predicted trajectories
        $\tau^{(k)}
        \gets F_{\mathrm{dyn}}(a^{(k)},s^q)$\;

        Evaluate the adversarial objective
        $\mathcal{L}_{\mathrm{adv}}
        (\tau^{(k)},A_c)$\;

        Evaluate the map-feasibility objective
        $\mathcal{L}_{\mathrm{map}}
        (\tau^{(k)},I_{\mathrm{map}},A_c)$\;

        Compute the guidance objective
        $\mathcal{L}_{\mathrm{guide}}
        \gets
        \lambda_{\mathrm{adv}}\mathcal{L}_{\mathrm{adv}}
        +
        \lambda_{\mathrm{map}}\mathcal{L}_{\mathrm{map}}$\;

        Mask the guidance gradient
        $g^{(k)}
        \gets
        M_{A_c}\odot
        \nabla_{u_{\mathrm{base}}^{(k)}}
        \mathcal{L}_{\mathrm{guide}}$\;

        Update the normalized action residual
        $u_{\mathrm{base}}^{(k+1)}
        \gets
        \operatorname{Adam}_{\gamma}
        \bigl(u_{\mathrm{base}}^{(k)},g^{(k)}\bigr)$\;
    }

    Reconstruct the final normalized actions
    $a_{\mathrm{n,final}}
    \gets
    \epsilon+u_{\mathrm{base}}^{(K_g)}$\;

    Map the final actions to physical units
    $a_{\mathrm{final}}
    \gets
    \mathcal{N}_a^{-1}(a_{\mathrm{n,final}})$\;

    Execute the first $L_e$ actions using
    $F_{\mathrm{dyn}}$\;

    Update $s^{q+1}$ and $H^{q+1}$\;

    Append the resulting states to $\mathcal{T}$\;
}

\KwRet{$\mathcal{T}$}
\end{algorithm}

\section{Experiments}

We evaluate RiskFlow on closed-loop safety-critical traffic scenario generation, focusing on both safety-critical controllability and behavioral realism.

\subsection{Experimental Setup}

We evaluate RiskFlow on nuScenes using TBSIM as the closed-loop simulation platform~\cite{caesar2020nuscenes,xu2023bits}. Models are trained on the nuScenes training split and evaluated on 100 validation scenes with diverse road topologies and interaction patterns.

We compare our method with representative closed-loop traffic generation methods, including STRIVE, BITS, CTG, CTG++, and CCDiff~\cite{rempe2022generating,xu2023bits,zhong2023guided,zhong2023language,lin2025causal}. All methods are evaluated on the same initial scenes under a consistent closed-loop evaluation protocol.

Table~\ref{tab:riskflow_hyperparameters} summarizes the evaluation protocol and the settings of RiskFlow
and the baselines. All methods use a temporal resolution of
$0.1$~s, 31 history steps, and 52 future steps, corresponding
to a $5.2$~s prediction horizon. RiskFlow represents each
future step as a six-dimensional state-action vector
$[x,y,v,\psi,a,\omega]$. In the default closed-loop setting,
the first $L_e=5$ generated actions are executed before
replanning. For the long-horizon experiments, the execution
interval is varied over $\{10,20,30,40,50\}$ steps. We evaluate
different intervention levels by controlling the top
$K\in\{2,3,4,5,10\}$ agents selected by the TTC-based
critical-agent ranking.

RiskFlow is trained for 100,000 optimization steps using Adam, a learning rate of $1\times10^{-4}$, a batch size of 4, and training seed 1. The MeanFlow time pair $(t,r)$ is sampled from a logit-normal distribution, with $r\neq t$ used for 80\% of the training samples. The JVP correction is scaled by $t-r$ and clipped to $[-0.03,0.03]$. At each planning cycle, RiskFlow performs one network evaluation for base action
generation, followed by $K_g=30$ output-space guidance updates. The adversarial and map-guidance weights are set to $50$ and $1$, respectively. The baselines use 100 reverse denoising steps, 30 guidance-gradient steps, and
collision/off-road guidance weights of $-50/1$. All methods are evaluated under the same controllable-agent and closed-loop planning configurations.

\begin{table*}[!t]
\caption{Hyperparameters of RiskFlow and the baselines used in
our experiments.}
\label{tab:riskflow_hyperparameters}
\centering
\small
\setlength{\tabcolsep}{5.5pt}
\begin{tabular}{l c l c}
\toprule
\textbf{Parameter Name} & \textbf{Value} &
\textbf{Parameter Name} & \textbf{Value} \\
\midrule

RiskFlow state-action dim. $D$ & 6 &
Step length & 0.1 s \\

History steps & 31 &
Generation steps & 52 \\

RiskFlow execution interval $L_e$ & 5 &
Long-horizon planning steps & 10, 20, 30, 40, 50 \\

Controllable agents & 2, 3, 4, 5, 10 &
Distance threshold & 50 m \\

Map encoder & ResNet-18 &
RiskFlow map feature dim. & 128 \\

Baseline map feature dim. & 256 &
Baseline trajectory encoder & MLP \\

Baseline trajectory feature dim. & 128 &
RiskFlow history/edge feature dim. & 64 \\

RiskFlow encoder/decoder layers & 2/2 &
Baseline Transformer dec. layers & 2 \\

RiskFlow attention heads & 16 &
Baseline Transformer dec. heads & 16 \\

RiskFlow hidden dim. & 384 &
RiskFlow edge-decoder layers & 4 \\

RiskFlow training LR & $1\times10^{-4}$ &
Baseline training LR & $1\times10^{-4}$ \\

RiskFlow batch size & 4 &
Baseline batch size & 100 \\

RiskFlow training steps & 100,000 &
Baseline training epochs & 10 \\

RiskFlow train/eval seeds & 1/0 &
Training optimizer & Adam \\

EMA update interval & 1 &
EMA decay & 0.995 \\

RiskFlow reconstruction weights $w_d$ & 1 for all $d$ &
Baseline trajectory-loss weight & 1.0 \\

RiskFlow JVP clamp & 0.03 &
Baseline yaw regularization weight & 0.1 \\

RiskFlow ratio of $r\neq t$ & 0.8 &
RiskFlow $(t,r)$ sampler & Logit-normal \\

RiskFlow JVP coefficient & $t-r$ &
RiskFlow map temporal decay & 0.9 \\

RiskFlow base-network evals. & 1 &
Baseline denoising steps & 100 \\

RiskFlow guidance steps $K_g$ & 30 &
Baseline guidance steps & 30 \\

RiskFlow guidance LR $\gamma$ & $1\times10^{-3}$ &
Baseline guidance LR & $1\times10^{-3}$ \\

RiskFlow guidance optimizer & Adam &
Baseline guidance optimizer & Adam \\

RiskFlow adversarial/map weights & $50/1$ &
Baseline collision/off-road weights & $-50/1$ \\

Baseline guidance constraint norm & 100 &
Baseline guidance discount factor & 0.99 \\

RiskFlow TTC clipping threshold & 20 s &
Baseline TTC threshold & 3.0 s \\

RiskFlow footprint samples & $10\times10$ &
Map feature temporal weighting & $0.9^{\ell-1}$ \\

\bottomrule
\end{tabular}
\end{table*}

\subsection{Evaluation Metrics}
\label{subsec:evaluation_metrics}

We report five raw metrics: Scenario Collision Rate (SCR), Scenario Off-Road Rate (ORR), Average Displacement Error (ADE), Final Displacement Error (FDE), and Kinematic Distribution Distance (KDD). Let $E$ denote the number of evaluated scenes, indexed by $n\in\{1,\ldots,E\}$. For scene
$n$, let $\mathcal{A}_n$ be its set of agents and
$\mathcal{V}_{n,i}$ be the valid rollout steps of agent $i$.
The generated and reference positions at step $t$ are denoted by $\hat p_{n,i}^{t}$ and $p_{n,i}^{t}$, respectively, with displacement error
$d_{n,i}^{t}=\lVert\hat p_{n,i}^{t}-p_{n,i}^{t}\rVert_2$.

\subsubsection{Average Displacement Error (ADE)}
ADE first averages the displacement error over the valid steps
of each agent and then over agents and scenes:
\begin{equation}
\mathrm{ADE}
=
\frac{1}{E}
\sum_{n=1}^{E}
\frac{1}{|\mathcal{A}_n|}
\sum_{i\in\mathcal{A}_n}
\frac{1}{|\mathcal{V}_{n,i}|}
\sum_{t\in\mathcal{V}_{n,i}}
d_{n,i}^{t}.
\label{eq:ade}
\end{equation}

\subsubsection{Final Displacement Error (FDE)}
Let $\hat p_{n,i}^{\mathrm{last}}$ and
$p_{n,i}^{\mathrm{last}}$ denote the generated and reference positions at the last valid step of agent $i$. FDE is defined as
\begin{equation}
\mathrm{FDE}
=
\frac{1}{E}
\sum_{n=1}^{E}
\frac{1}{|\mathcal{A}_n|}
\sum_{i\in\mathcal{A}_n}
d_{n,i}^{\mathrm{last}}.
\label{eq:fde}
\end{equation}

\subsubsection{Scenario Collision Rate}
Let $c_n\in\{0,1\}$ indicate whether any pair of agents collides during the rollout of scene $n$. SCR is the fraction of scenes with at least one collision:
\begin{equation}
\mathrm{SCR}
=
\frac{1}{E}
\sum_{n=1}^{E}c_n.
\label{eq:scr}
\end{equation}

\subsubsection{Scenario Off-Road Rate}
Let $\mathcal{D}_n$ denote the drivable region of scene $n$,
and let
$\Omega_n=\{(i,t)\mid i\in\mathcal{A}_n,\,
t\in\mathcal{V}_{n,i}\}$ contain all valid agent-time pairs.
ORR computes the percentage of valid states outside the drivable region for each scene and then averages it across scenes:
\begin{equation}
\mathrm{ORR}
=
\frac{100}{E}
\sum_{n=1}^{E}
\frac{1}{|\Omega_n|}
\sum_{(i,t)\in\Omega_n}
\mathbf{1}
\left[
\hat p_{n,i}^{t}\notin\mathcal{D}_n
\right].
\label{eq:orr}
\end{equation}

\subsubsection{Kinematic Distribution Distance}
KDD measures the discrepancy between generated and
human-driving distributions of longitudinal acceleration, lateral acceleration, and jerk. Let
$\mathcal{K}=\{a_{\mathrm{lon}},a_{\mathrm{lat}},j\}$. For
each $k\in\mathcal{K}$, let $P_{\mathrm{gt}}^{k}$ and $P_{\mathrm{gen}}^{k}$ denote the normalized ground-truth and generated histograms. Their Earth Mover's Distance is
\begin{equation}
D_k
=
\operatorname{EMD}
\left(
P_{\mathrm{gt}}^{k},
P_{\mathrm{gen}}^{k}
\right),
\label{eq:kdd_component}
\end{equation}
where the transport cost is determined by the absolute distance between histogram-bin values. KDD is the average distance over the three kinematic variables:
\begin{equation}
\mathrm{KDD}
=
\frac{1}{|\mathcal{K}|}
\sum_{k\in\mathcal{K}}D_k.
\label{eq:kdd}
\end{equation}

Higher SCR indicates stronger safety-critical controllability, whereas lower ORR, ADE, FDE, and KDD indicate better realism. Following CCDiff~\cite{lin2025causal}, we min-max normalize the SCR of method $p$ to obtain the
Controllability Score:
\begin{equation}
\mathrm{CS}_p
=
\frac{
\mathrm{SCR}_p-\min(\mathrm{SCR})
}{
\max(\mathrm{SCR})-\min(\mathrm{SCR})
}.
\label{eq:cs}
\end{equation}

Let
$\mathcal{E}
=\{\mathrm{ORR},\mathrm{ADE},\mathrm{FDE},\mathrm{KDD}\}$.
The Realism Score is defined as
\begin{equation}
\mathrm{RS}_p
=
1-\frac{1}{4}
\sum_{m\in\mathcal{E}}
\frac{
m_p-\min(m)
}{
\max(m)-\min(m)
}.
\label{eq:rs}
\end{equation}

The same extrema are shared across all experiments, and each normalized term is clipped to $[0,1]$. Higher CS and RS indicate better controllability and realism, respectively.

\begin{figure*}[!t]
    \centering
    \includegraphics[width=0.60\textwidth]{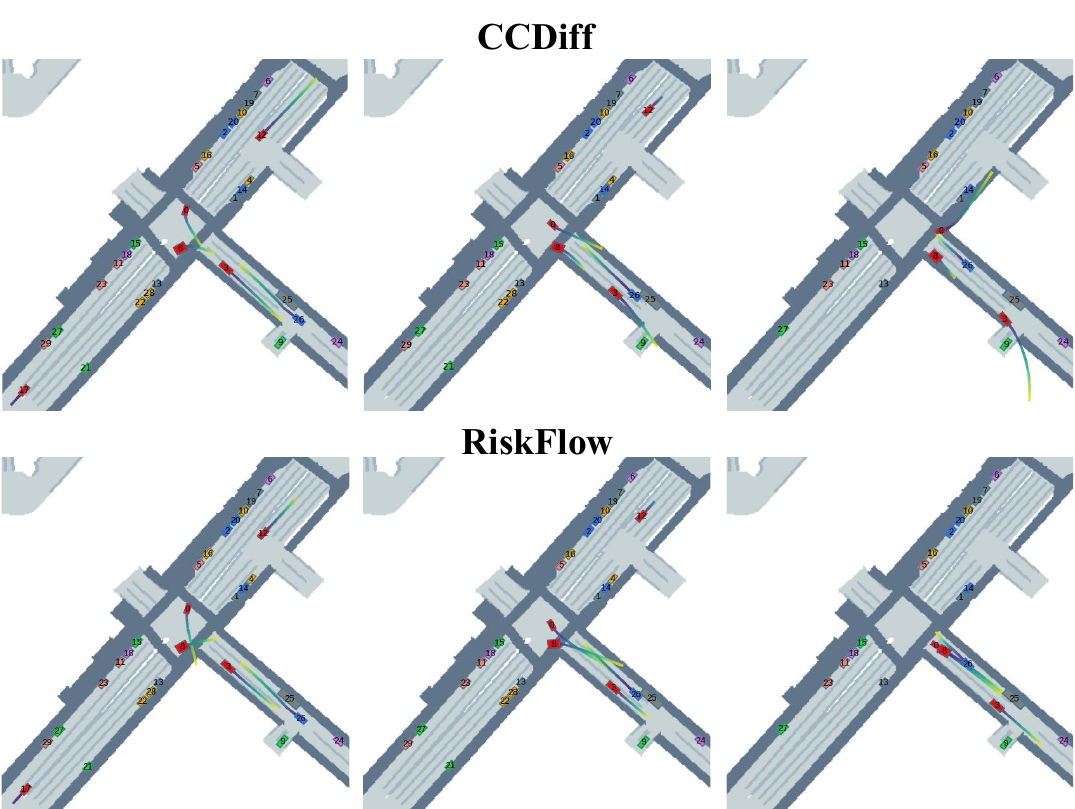}
    \caption{Qualitative comparison on Scene 0556 with $K=5$ and $T=0.5$ s.
    CCDiff fails to generate a valid collision, while RiskFlow successfully
    induces a road-consistent collision.}
    \label{fig:scene_0556}
\end{figure*}

\subsection{Multi-Agent Scenario Generation}

We evaluate multi-agent scenario generation by varying the number of controllable agents over $K\in\{2,3,4,5,10\}$.
Table~\ref{tab:multi_agent_cs_rs} reports the corresponding CS and RS. As the number of controllable agents increases, preserving realistic behavior becomes increasingly challenging because stronger interventions can introduce off-road motion and larger trajectory deviations during closed-loop scenario execution.

STRIVE achieves the highest CS across all settings, reaching 1.00 at $K=10$. However, its RS decreases substantially from 0.62 at $K=2$ to 0.14 at $K=10$, indicating that its strong collision-generation capability comes at the cost of degraded realism. Other baselines exhibit a similar trade-off. For example, CCDiff obtains the second-best CS at $K=5$ and $K=10$, while its RS decreases from 0.75 to 0.39 as the number of controlled agents increases.

In contrast, RiskFlow achieves the best RS across all control settings, with scores of 0.92, 0.85, 0.78, 0.74, and 0.58 for
$K=2,3,4,5,$ and $10$, respectively. Although RiskFlow is not the most aggressive collision generator, it ties for the second-best CS at $K=5$ and provides substantially stronger road feasibility, spatial accuracy, and kinematic realism. These results demonstrate a balance between safety-critical controllability and realistic multi-agent behavior.

\begin{table}[t]
\caption{Comparison of Controllability Score (CS) and Realism Score (RS) for multi-agent scenario generation under different numbers of controllable agents. Best results are shown in bold, and second-best results are underlined.}
\label{tab:multi_agent_cs_rs}
\centering
{
\setlength{\tabcolsep}{2pt}
\begin{tabular}{lccccc}
\toprule
Method & $K=2$ & $K=3$ & $K=4$ & $K=5$ & $K=10$ \\
\midrule

\multicolumn{6}{c}{\textbf{Controllability Score (CS)}} \\
\midrule

STRIVE
& \textbf{0.68}
& \textbf{0.68}
& \textbf{0.86}
& \textbf{0.86}
& \textbf{1.00} \\

BITS
& 0.18
& 0.18
& 0.14
& 0.23
& 0.14 \\

CTG
& \underline{0.41}
& 0.36
& \underline{0.55}
& 0.36
& 0.45 \\

CTG++
& 0.36
& \underline{0.55}
& 0.45
& 0.45
& 0.32 \\

CCDiff
& 0.27
& 0.45
& 0.41
& \underline{0.55}
& \underline{0.68} \\

Ours
& 0.23
& 0.27
& 0.41
& \underline{0.55}
& 0.32 \\

\midrule
\multicolumn{6}{c}{\textbf{Realism Score (RS)}} \\
\midrule

STRIVE
& 0.62
& 0.48
& 0.38
& 0.32
& 0.14 \\

BITS
& 0.77
& 0.68
& 0.65
& 0.57
& \underline{0.53} \\

CTG
& 0.78
& 0.72
& \underline{0.67}
& \underline{0.64}
& 0.48 \\

CTG++
& \underline{0.82}
& \underline{0.74}
& 0.66
& 0.62
& 0.51 \\

CCDiff
& 0.75
& 0.64
& 0.59
& 0.54
& 0.39 \\

Ours
& \textbf{0.92}
& \textbf{0.85}
& \textbf{0.78}
& \textbf{0.74}
& \textbf{0.58} \\

\bottomrule
\end{tabular}
}
\end{table}

\subsection{Long-Horizon Closed-Loop Generation}

We further evaluate closed-loop generation with execution horizons from $T=1$~s to $T=5$~s. Table~\ref{tab:long_horizon_cs_rs} reports the corresponding CS and RS. A longer execution horizon reduces the replanning frequency and allows prediction and guidance errors to
propagate for more steps, increasing trajectory drift, off-road motion, and abnormal kinematic behavior.

RiskFlow achieves the highest RS at $T=1$~s and $T=2$~s, with scores of 0.74 and 0.71, respectively. It obtains the second-best RS at $T=3$--$5$~s, with scores of 0.56, 0.48, and 0.54. At $T=3$~s, RiskFlow and CTG++ both round to 0.56, but CTG++ ranks first according to the unrounded scores. In comparison, STRIVE exhibits substantial realism degradation as the horizon increases, with its RS decreasing from 0.47 at $T=1$~s to 0.30 at $T=5$~s.

RiskFlow also maintains competitive controllability across different execution horizons. It achieves the best CS of 0.55 at $T=2$~s and ties with CCDiff for the best CS of 0.82 at $T=5$~s. It additionally obtains the second-best CS of 0.55 at $T=1$~s. CTG and CCDiff achieve higher CS at $T=3$~s and $T=4$~s, whereas RiskFlow preserves stronger realism than both methods at these horizons. Overall, these results show that RiskFlow maintains an effective balance between safety-critical
event generation and behavioral realism over extended closed-loop rollouts.

\begin{table}[t]
\caption{Comparison of Controllability Score (CS) and Realism Score (RS) for long-horizon closed-loop generation under different execution horizons $T$. Best results are shown in bold, and second-best results are underlined.}
\label{tab:long_horizon_cs_rs}
\centering
{
\setlength{\tabcolsep}{2pt}
\begin{tabular}{lccccc}
\toprule
Method & $T=1$ s & $T=2$ s & $T=3$ s & $T=4$ s & $T=5$ s \\
\midrule

\multicolumn{6}{c}{\textbf{Controllability Score (CS)}} \\
\midrule

STRIVE
& \textbf{0.68}
& 0.32
& 0.32
& 0.23
& 0.18 \\

BITS
& 0.14
& 0.00
& 0.45
& 0.23
& 0.32 \\

CTG
& 0.32
& \underline{0.45}
& \textbf{0.68}
& \underline{0.82}
& \underline{0.68} \\

CTG++
& 0.41
& 0.41
& \underline{0.64}
& 0.45
& \underline{0.68} \\

CCDiff
& 0.27
& \underline{0.45}
& \textbf{0.68}
& \textbf{0.95}
& \textbf{0.82} \\

Ours
& \underline{0.55}
& \textbf{0.55}
& 0.55
& 0.73
& \textbf{0.82} \\

\midrule
\multicolumn{6}{c}{\textbf{Realism Score (RS)}} \\
\midrule

STRIVE
& 0.47
& 0.48
& 0.32
& 0.24
& 0.30 \\

BITS
& 0.59
& 0.57
& 0.49
& 0.47
& 0.47 \\

CTG
& 0.62
& 0.59
& 0.38
& 0.29
& 0.48 \\

CTG++
& \underline{0.70}
& \underline{0.70}
& \textbf{0.56}
& \textbf{0.51}
& \textbf{0.63} \\

CCDiff
& 0.62
& 0.66
& 0.51
& 0.46
& 0.52 \\

Ours
& \textbf{0.74}
& \textbf{0.71}
& \underline{0.56}
& \underline{0.48}
& \underline{0.54} \\

\bottomrule
\end{tabular}
}
\end{table}

\begin{table}[t]
\caption{Inference speed comparison under $K=5$ and $T=0.5$ s on 100 scenes using a single NVIDIA RTX 4090 GPU under identical evaluation settings. Best results are shown in bold, and second-best results are underlined.}
\label{tab:inference_speed}
\centering
{
\small
\setlength{\tabcolsep}{4pt}
\begin{tabular}{lccc}
\toprule
Method & Total Time & Avg. Time per Scene & Runtime Ratio \\
\midrule
CTG++
& 30.27 h
& 18.16 min
& 22.42\(\times\) \\

CCDiff
& 6.25 h
& 3.75 min
& 4.63\(\times\) \\

CTG
& \underline{2.77 h}
& \underline{1.66 min}
& \underline{2.05\(\times\)} \\

Ours
& \textbf{1.35 h}
& \textbf{48.6 s}
& \textbf{1.00\(\times\)} \\
\bottomrule
\end{tabular}
}
\end{table}

\subsection{Qualitative Comparison}
Fig.~\ref{fig:scene_0556} compares CCDiff and RiskFlow on Scene 0556 with $K=5$ controllable agents and an execution horizon of $T=0.5$ s. CCDiff fails to induce a collision, and its guided trajectory leaves the drivable region, revealing unrealistic artifacts from iterative closed-loop guidance.

In contrast, RiskFlow induces a road-aligned collision with smooth, physically plausible, and temporally coherent trajectories throughout the closed-loop rollout, avoiding abrupt lateral shifts or implausible controls. This example demonstrates its ability to generate safety-critical events while preserving road feasibility and behavioral realism.

\subsection{Inference Speed}

Table~\ref{tab:inference_speed} compares the inference efficiency of the evaluated methods under the same setting, using $K=5$ controllable agents, an execution horizon of $T=0.5$~s, and a single NVIDIA RTX 4090 GPU. We measure the end-to-end wall-clock time required to complete the closed-loop evaluation of 100 scenes and obtain the average runtime per scene by dividing the total time by 100. For a fair comparison, all methods use the same guidance budget of 30 Adam optimization steps at each planning cycle.

RiskFlow completes the evaluation in 1.35 hours, corresponding to an average runtime of 48.6 seconds per scene. In comparison, CTG++, CCDiff, and CTG require 30.27, 6.25, and 2.77 hours, respectively. RiskFlow therefore achieves $22.42\times$, $4.63\times$, and $2.05\times$ speedups over these methods.

Since the guidance optimizer and the number of guidance steps are kept consistent, the efficiency improvement mainly comes from the base-generation stage. RiskFlow requires only one MeanFlow network evaluation (1 NFE) to generate the base action sequence, whereas the baselines use 100 reverse-denoising network evaluations.

\subsection{Ablation Study}

Table~\ref{tab:ablation_results} reports the raw metrics used to isolate the effects of MeanFlow generation, adversarial guidance $\mathcal{L}_{\mathrm{adv}}$, and map guidance
$\mathcal{L}_{\mathrm{map}}$ under different numbers of controllable agents. An empty entry in the MF column denotes the diffusion backbone, while the final row corresponds to the complete RiskFlow configuration.

\begin{table*}[!t]
\caption{Detailed ablation results under different numbers of controllable agents. D, MF, A, and M denote diffusion, MeanFlow, adversarial guidance, and map guidance, respectively. RiskFlow uses MF, A, and M. Best and second-best results for each metric and $K$ are shown in bold and underlined, respectively.}
\label{tab:ablation_results}
\centering
\footnotesize
\setlength{\tabcolsep}{5.5pt}
\renewcommand{\arraystretch}{0.92}
\begin{tabular}{clcccccccc}
\toprule
$K$ & Metric
& D
& D+A
& D+M
& D+A+M
& MF
& MF+A
& MF+M
& RiskFlow \\
\midrule

\multirow{5}{*}{$2$}
& SCR $\uparrow$
& \textbf{0.46}
& 0.33
& 0.40
& 0.40
& \underline{0.44}
& 0.40
& 0.38
& 0.39 \\
& ORR $\downarrow$
& 2.84
& 2.70
& 1.31
& \underline{0.61}
& 1.45
& 1.49
& 0.68
& \textbf{0.44} \\
& FDE $\downarrow$
& 4.39
& 4.45
& 4.20
& 4.17
& 3.20
& 3.21
& \textbf{3.05}
& \underline{3.14} \\
& KDD $\downarrow$
& 10.00
& 10.26
& 10.30
& 9.90
& 6.45
& \textbf{6.29}
& \underline{6.34}
& 6.68 \\
& ADE $\downarrow$
& 1.61
& 1.67
& 1.55
& 1.55
& 1.26
& \underline{1.25}
& \textbf{1.20}
& \underline{1.25} \\
\midrule

\multirow{5}{*}{$3$}
& SCR $\uparrow$
& \underline{0.49}
& 0.36
& 0.44
& 0.44
& \textbf{0.50}
& 0.41
& 0.39
& 0.40 \\
& ORR $\downarrow$
& 3.08
& 3.04
& 1.48
& 0.72
& 1.47
& 1.53
& \underline{0.69}
& \textbf{0.48} \\
& FDE $\downarrow$
& 5.35
& 5.36
& 5.25
& 5.22
& \underline{4.03}
& 4.14
& \textbf{3.94}
& 4.15 \\
& KDD $\downarrow$
& 10.56
& 10.86
& 11.67
& 11.22
& \underline{6.70}
& \textbf{6.69}
& 6.77
& 6.72 \\
& ADE $\downarrow$
& 1.96
& 2.00
& 1.97
& 1.96
& \underline{1.59}
& 1.62
& \textbf{1.56}
& 1.64 \\
\midrule

\multirow{5}{*}{$4$}
& SCR $\uparrow$
& \underline{0.48}
& 0.36
& 0.40
& 0.43
& \textbf{0.51}
& 0.45
& 0.42
& 0.43 \\
& ORR $\downarrow$
& 3.95
& 3.81
& 1.60
& \underline{0.99}
& 1.86
& 1.88
& 1.11
& \textbf{0.95} \\
& FDE $\downarrow$
& 6.47
& 6.48
& 6.00
& 5.99
& 4.78
& 4.84
& \textbf{4.63}
& \underline{4.70} \\
& KDD $\downarrow$
& 11.09
& 11.21
& 11.46
& 10.95
& 7.28
& \underline{7.16}
& \textbf{7.09}
& 7.45 \\
& ADE $\downarrow$
& 2.33
& 2.34
& 2.24
& 2.24
& 1.88
& 1.88
& \textbf{1.80}
& \underline{1.85} \\
\midrule

\multirow{5}{*}{$5$}
& SCR $\uparrow$
& \underline{0.48}
& 0.33
& 0.42
& 0.46
& \textbf{0.49}
& 0.45
& 0.44
& 0.46 \\
& ORR $\downarrow$
& 4.25
& 4.12
& 2.01
& \underline{1.02}
& 2.03
& 2.06
& \textbf{0.89}
& \textbf{0.89} \\
& FDE $\downarrow$
& 7.14
& 7.16
& 6.71
& 6.59
& \underline{5.28}
& 5.35
& \textbf{5.19}
& \textbf{5.19} \\
& KDD $\downarrow$
& 10.94
& 11.40
& 11.54
& 11.08
& 7.54
& \textbf{7.35}
& \underline{7.39}
& 7.69 \\
& ADE $\downarrow$
& 2.59
& 2.63
& 2.53
& 2.53
& 2.10
& 2.10
& \underline{2.06}
& \textbf{2.05} \\
\midrule

\multirow{5}{*}{$10$}
& SCR $\uparrow$
& \underline{0.48}
& 0.35
& 0.44
& \textbf{0.49}
& 0.47
& 0.41
& 0.40
& 0.41 \\
& ORR $\downarrow$
& 5.19
& 4.86
& 2.23
& 1.80
& 2.32
& 2.39
& \underline{1.03}
& \textbf{1.01} \\
& FDE $\downarrow$
& 8.90
& 8.91
& 8.66
& 7.84
& \underline{6.72}
& 6.81
& \textbf{6.65}
& 6.73 \\
& KDD $\downarrow$
& 11.81
& 12.29
& 12.17
& 11.82
& 9.28
& \underline{9.18}
& \textbf{9.01}
& 9.55 \\
& ADE $\downarrow$
& 3.26
& 3.26
& 3.27
& 3.23
& 2.67
& 2.68
& \textbf{2.63}
& \underline{2.66} \\
\bottomrule
\end{tabular}
\end{table*}

Replacing diffusion with MeanFlow consistently improves trajectory realism and physical consistency. Compared with D, MF substantially reduces ORR, FDE, KDD, and ADE across all values of $K$, while maintaining a comparable SCR. At $K=5$, for example, MF reduces ORR from 4.25 to 2.03, FDE from 7.14 to 5.28, KDD from 10.94 to 7.54, and ADE from 2.59 to 2.10. The improvement becomes more evident as the number of controllable agents increases. In particular, the FDE reduction increases from 1.19 at $K=2$ to 2.18 at $K=10$, indicating that one-step action generation remains effective in increasingly complex multi-agent interactions.

Map guidance mainly improves road feasibility. Adding M reduces ORR for both generators at every value of $K$. For instance, at $K=10$, it reduces ORR from 5.19 to 2.23 for diffusion and from 2.32 to 1.03 for MeanFlow. It also generally improves FDE and ADE, suggesting that enforcing map constraints suppresses off-road behavior without sacrificing trajectory realism. The consistent reduction in ORR becomes especially important for larger $K$, where accumulated interactions make road-compliant generation more difficult.

Finally, adding adversarial guidance to MF+M produces the complete RiskFlow model. Compared with MF+M, RiskFlow improves SCR from 0.38 to 0.39, 0.39 to 0.40, 0.42 to 0.43, 0.44 to 0.46, and 0.40 to 0.41 for $K=2,3,4,5,$ and $10$, respectively. Meanwhile, ORR is either reduced or unchanged. At $K=5$, RiskFlow increases SCR to 0.46 while retaining the same ORR and FDE as MF+M and slightly reducing ADE from 2.06 to 2.05. Although adversarial guidance introduces small variations in FDE, ADE, and KDD in several settings, the overall realism remains competitive. These results demonstrate that MeanFlow provides an effective generative prior, map guidance preserves road feasibility, and adversarial guidance strengthens safety-critical generation while maintaining realistic closed-loop behavior.

\section{Conclusion}

We introduce \textbf{RiskFlow}, an efficient closed-loop framework for safety-critical traffic scenario generation. RiskFlow transforms Gaussian action sequences into future acceleration and yaw-rate sequences through one MeanFlow pass and applies test-time guidance directly in action space. TTC-based critical-agent selection, vehicle-dynamics rollout, and map-aware regularization enable adversarial interactions while preserving road feasibility and physical plausibility. Experiments on nuScenes demonstrate a favorable adversariality-realism trade-off across multi-agent and long-horizon settings, with substantially improved inference efficiency. Ablation studies confirm that MeanFlow improves
trajectory realism and map guidance stabilizes interventions under road constraints. Overall, RiskFlow provides an efficient, physically grounded approach to closed-loop safety-critical traffic generation.

\bibliographystyle{IEEEtran}
\bibliography{IEEEabrv,riskflow}

\end{document}